\documentclass{article} 
\usepackage{iclr2024_conference,times}

\usepackage{amsmath,amsfonts,bm}

\def\eqref#1{equation~\ref{#1}}

\def\1{\bm{1}}

\DeclareMathAlphabet{\mathsfit}{\encodingdefault}{\sfdefault}{m}{sl}
\SetMathAlphabet{\mathsfit}{bold}{\encodingdefault}{\sfdefault}{bx}{n}

\usepackage{hyperref}
\usepackage{url}
\usepackage{graphicx}
\usepackage{wrapfig}
\usepackage{subcaption}

\title{Faraday: Synthetic Smart Meter Generator for the smart grid}

\author{Sheng Chai \&  Gus Chadney \\
Centre for Net Zero\\
\texttt{\{sheng.chai,gus.chadney\}@centrefornetzero.org} \\
}

\iclrfinalcopy 
\begin{document}

\maketitle

\begin{abstract}
Access to smart meter data is essential to rapid and successful transitions to electrified grids, underpinned by flexibility delivered by low carbon technologies, such as  electric vehicles (EV) and heat pumps, and powered by renewable energy. Yet little of this data is available for research and modelling purposes due consumer privacy protections. Whilst many are calling for raw datasets to be unlocked through regulatory changes, we believe this approach will take too long. Synthetic data addresses these challenges directly by overcoming privacy issues. In this paper, we present Faraday, a Variational Auto-encoder (VAE)-based model trained over 300 million smart meter data readings from an energy supplier in the UK, with information such as property type and low carbon technologies (LCTs) ownership. The model produces household-level synthetic load profiles conditioned on these labels, and we compare its outputs against actual substation readings to show how the model can be used for real-world applications by grid modellers interested in modelling energy grids of the future.

\end{abstract}

\section{Introduction}

A huge part of the global transition to net zero involves increasing the share of electricity from renewable sources and concomitantly electrifying heating and transport. Given the variability of renewables and growing adoption of low carbon technologies (LCTs), there are new challenges to our electricity grid such as creating new demand peaks, grid constraints and mismatches between demand and supply. Households with LCTs however can provide flexibility through automation to solve these challenges \citep{ng_2021, ofgem_ev_priorities, xilas_britain_2023, cel_grid_2022}. 

With access to granular household-level electricity consumption data, especially of households with LCTs, we can build better bottom-up grid models of future energy systems to model scenarios, such as varying frequencies of adverse weather conditions or different heat pump adoption rates to understand grid resiliency, identify grid constraints, and plan for grid reinforcement projects \citep{damianakis_assessing_2023, chen_control_2023}. Many countries have rolled out smart meters to collect electricity consumption at 15 or 30-minute intervals. Access to this data is limited due to privacy concerns. An example of a smart meter data repository in the United Kingdom is the Smart Energy Research Lab \citep{SERL} which has a strict approval criteria. Calls to liberalise smart meter data for public good \citep{esc_dcc} are growing, but they too involve lengthy bureaucracy and strict governance processes. Generating synthetic smart meter data circumnavigates these issues. 

There is a growing number of literature \citep{zhang_generative_2019, liang_synthesis_2022, gu_ganb_2019} that have applied generative artificial intelligence technologies to create synthetic smart meter data. Yet many of them are only theoretical and lack the ability to condition outputs on the type of LCTs that households own. Understanding how different households with LCTs consume electricity is  important in modelling grid systems of the future.
 
In this paper, we present Faraday - a model trained using a combination of Variational Auto-encoder and Gaussian Mixture Model. We demonstrate how it has been evaluated and tested in the real world. The model is trained on a proprietary dataset of UK households with metadata on the type of LCTs they own and as such is capable of conditioning outputs by this information. The model is currently deployed live as a web-app and an API in closed-alpha phase, available to about 50 alpha testers from academia and industry. 

\section{Methodology}

\subsection{About the dataset}

We have access to proprietary data belonging to an energy supplier in the United Kingdom. The dataset consists of metered half-hourly electricity consumption data of 20 thousand households in 2021 and 2022, alongside attributes such as the type of LCTs they own, the property type and the property's energy rating. The total number of smart meter readings in this dataset is in the excess of 300 million. 

The access to this rich dataset means Faraday is able to generate household-level synthetic smart meter profiles that are conditioned upon inputs that the user provides, such as whether the household owns an electric vehicle (EV), their capacity to shift around consumption in response to grid conditions (i.e. whether they have a smart tariff) and the type of property (e.g. bungalows, terraced or flats). The model’s outputs are valuable to grid modellers who want to model future energy systems that rely on low carbon technologies and renewable energy sources.

\subsection{Faraday architecture}

\begin{wrapfigure}{r}{0.5\textwidth}
  \begin{center}
    \includegraphics[width=0.48\textwidth]{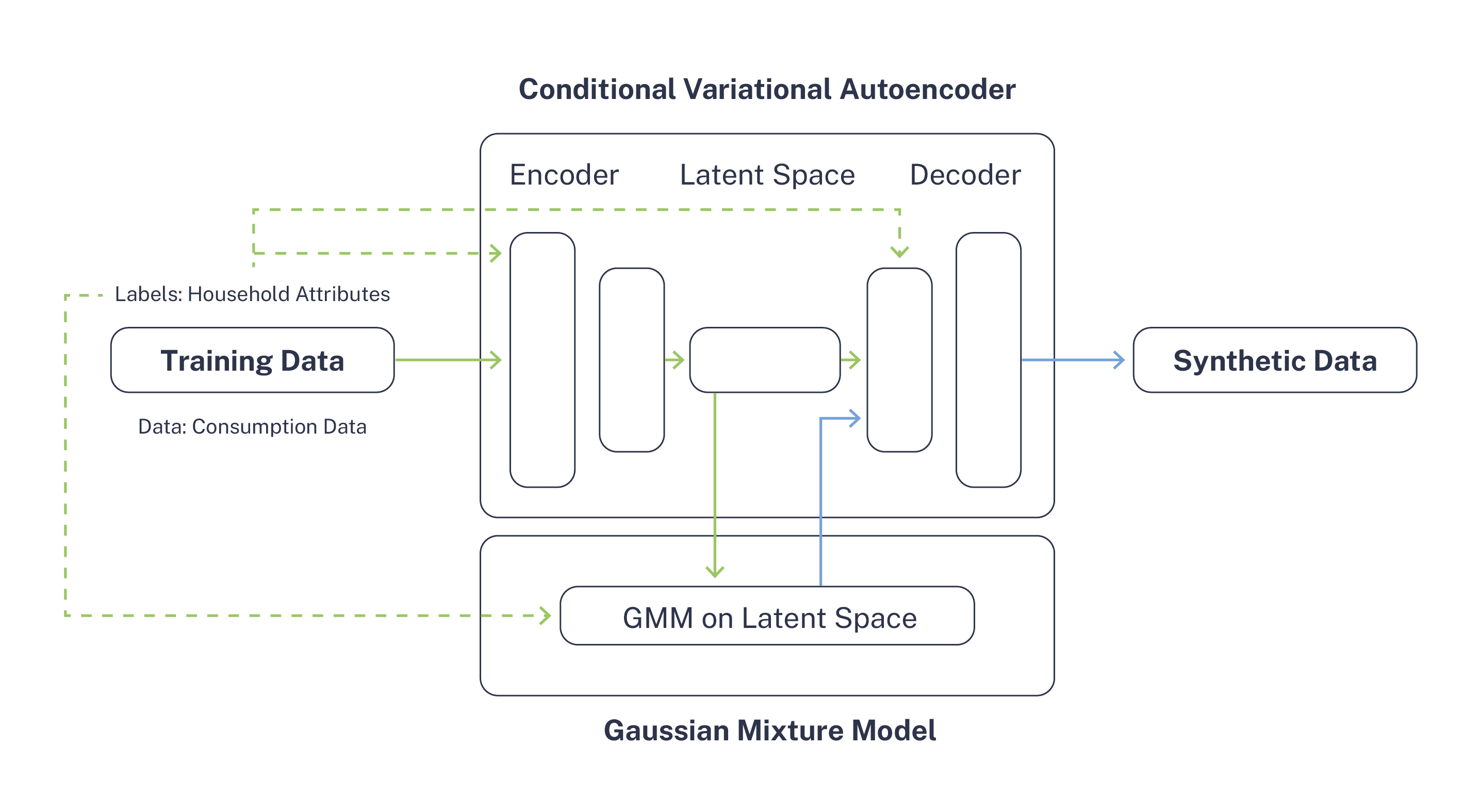}
  \end{center}
  \caption{Faraday architecture}
\end{wrapfigure}

Faraday is a model based on Conditional Variational Auto-encoder (VAE) \citep{cvae} and Gaussian Mixture Model (GMM) algorithms. It works by first training a conditional VAE. The training data is then mapped to the latent space using the trained auto-encoder on which a GMM is trained to learn the distribution of the latent space. During inference, a random sample of latent vectors is drawn from the GMM and decoded using the trained decoder. To support conditional sampling, labels are appended to the latent codes when training the GMM. During sampling, random samples are drawn and labels that do not match the user's inputs are discarded.

\subsubsection{Modifications to VAE}

A traditional VAE's reparametrization layer uses a normal distribution via the Kullback-Leibler (KL) Divergence loss to approximate the latent space. As smart meter data is non-normal and heavily positively skewed, KL-Divergence loss is not ineffective for learning this distribution. Applying log-normal transformation to the input data helps but comes at the cost of the fidelity at the higher quantiles (the peak loads) which is an area of interest when it comes to grid modelling. Instead of KL-Divergence loss, the Maximum Mean Divergence (MMD) loss is used in similar fashion to InfoVAE \citep{zhao_infovae_2018}. In addition, quantile losses at 5$^{th}$, 50$^{th}$ and 95$^{th}$ are added to the loss function to help improve fidelity of synthetic outputs at various quantiles. The loss function used is therefore the sum of: a. the reconstruction loss (mean-squared error), b. the MMD (instead of KL-Divergence) loss and c. the three quantile losses.

\subsubsection{Gaussian sampling of latent space}
The latent space of a traditional VAE is modelled using a unimodal Gaussian distribution. However, this is insufficient to capture the distribution of the latent space. To guarantee that the distribution of the generated samples matches that of the real data, a large random sample of the training data was drawn and encoded to the latent space with the trained encoder. A GMM is then trained to learn the distribution of the latent space where the training data occupies. During inference, random samples are drawn using the GMM and samples are decoded using the trained decoder. This is different to the original GM-VAE \citep{gmvae} where the Gaussian Mixture Model replaces the prior distribution of a unimodal Gaussian to learn the distribution of the latent space implicitly.

\section{Results}

In this section we analyse the outputs of Faraday through these three qualities \cite{jordon_synthetic_2022}: 1) Fidelity 2) Utility and 3) Privacy. We also compare the results to other GAN-based models for utility and demonstrate that Faraday outputs were of higher utility than GAN-based outputs.

\subsection{Fidelity}

\begin{figure}[t]
\includegraphics[width=1\textwidth]{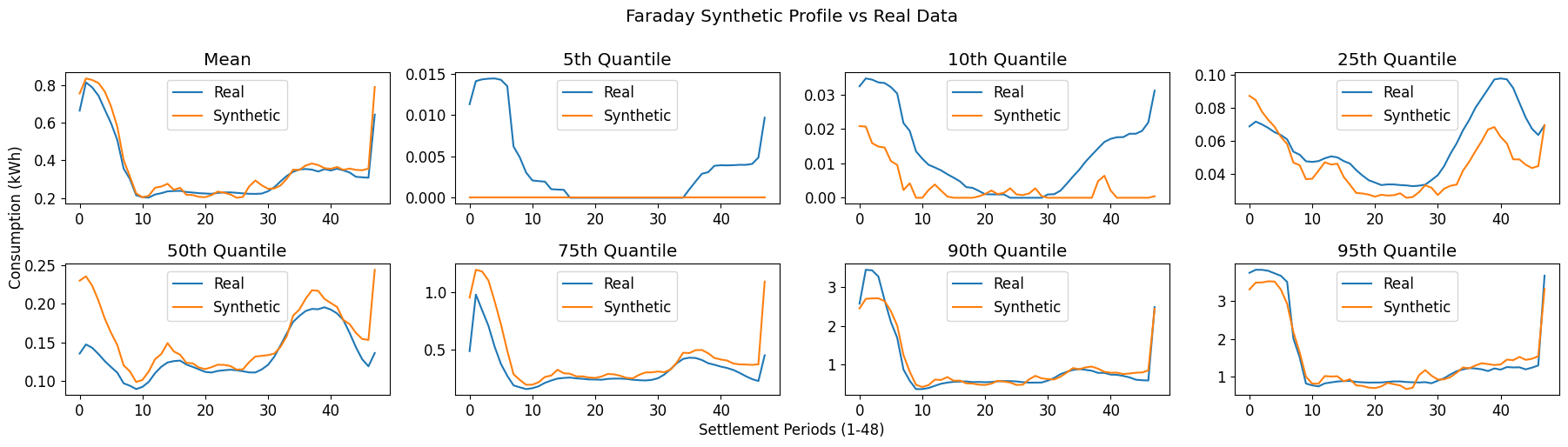}
\centering
\caption{Faraday outputs at various quantiles. Y-axis is the kWh consumption. X-axis is the half-hourly periods where 1 is 00:00hrs and 48 is 23:30 hrs.}
\label{fig:gmm_quantiles}
\end{figure}

\begin{figure}[ht]
\centering
\begin{subfigure}{0.49\textwidth}
\includegraphics[width=0.95\linewidth]{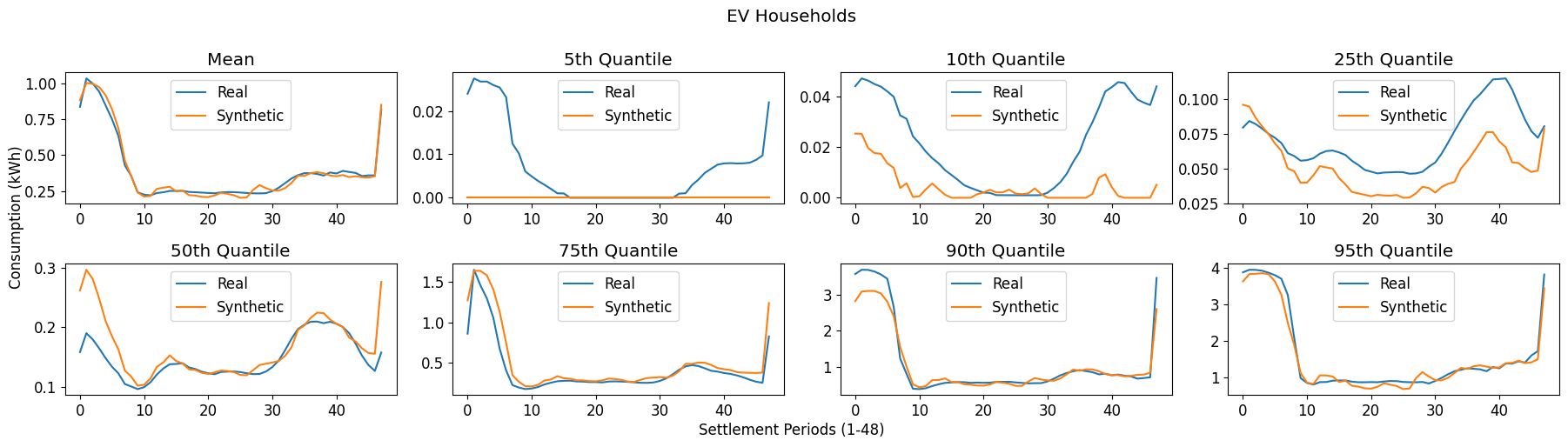} 
\caption{Load profile at various quantiles for Households with EV.}
\label{fig:ev_quantile}

\end{subfigure}
\begin{subfigure}{0.49\textwidth}
\includegraphics[width=0.95\linewidth]{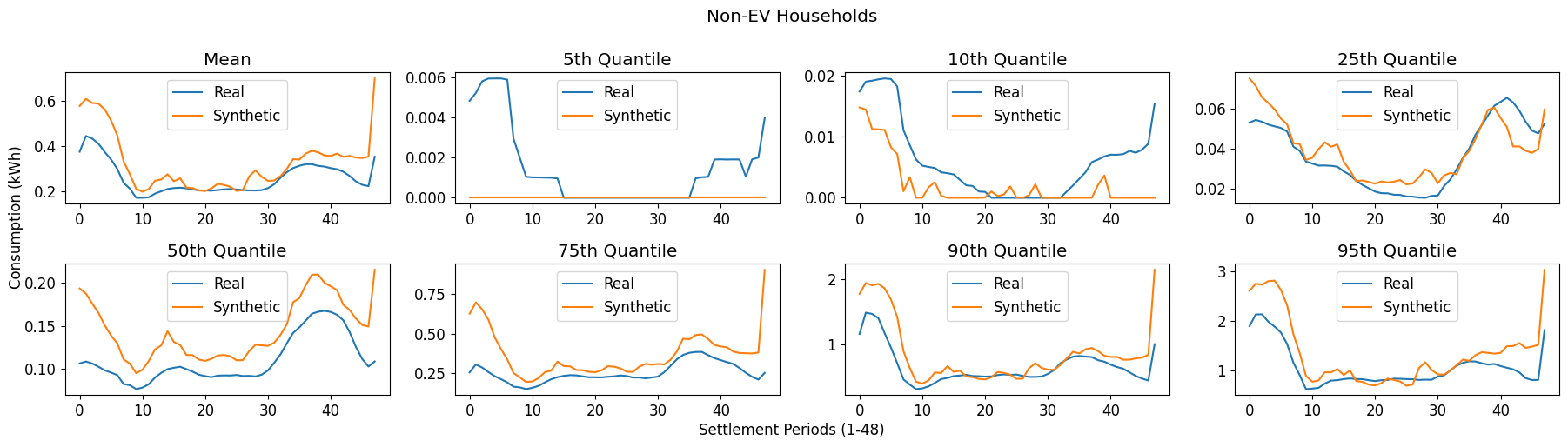}
\caption{Load profile at various quantiles for Households without EV.}
\label{fig:nonev_quantile}

\end{subfigure}
\caption{Faraday outputs at various quantiles conditioned by whether households own an electric vehicle (EV).}
\end{figure}

\begin{figure}[ht]
\centering
\begin{subfigure}{0.45\textwidth}
\includegraphics[width=0.9\linewidth]{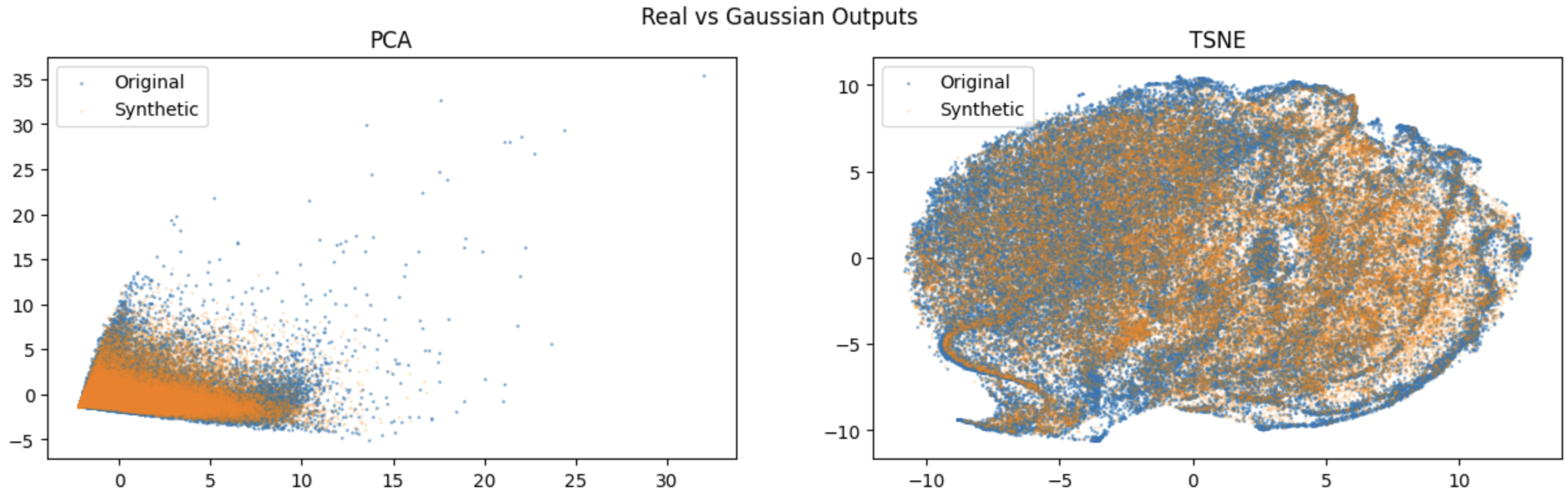} 
\caption{Distribution of Faraday outputs visualised with PCA and T-SNE plots.}
\label{fig:gmm_tsne}
\end{subfigure}
\begin{subfigure}{0.45\textwidth}
\includegraphics[width=0.9\linewidth]{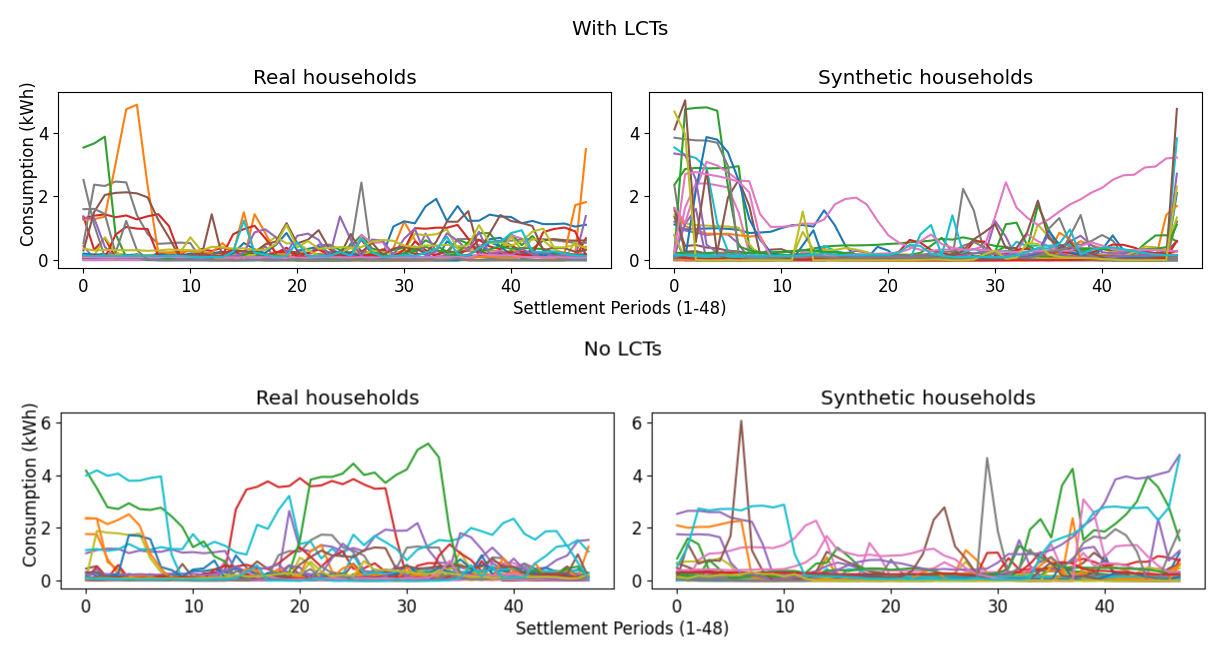}
\caption{Comparison between real vs Faraday outputs at individual household level.}
\label{fig:ev_no_ev_indiv}
\end{subfigure}
\caption{Fidelity of Faraday outputs.}
\end{figure}

Fidelity refers to the statistical similarity between synthetic and real data. This can be done quantitatively by comparing the statistical metrics (such as mean and quantile values), or qualitatively by comparing the distributions visually via t-stochastic neighbour embedding (TSNE) or principal component analysis (PCA) plots \citep{timegan}. Figures \ref{fig:gmm_quantiles} and \ref{fig:gmm_tsne} show high fidelity of Faraday outputs except at the 5$^{th}$ quantile where synthetic data is clipped to a minimum of zero. More tuning could be done to improve performance at the 5$^{th}$ quantile but comes at the expense of performance at the 95$^{th}$ quantile. Due to grid modellers’ interest in studying peak loads, performance at higher quantiles is prioritised over performance at lower quantiles. Figures \ref{fig:ev_quantile} and \ref{fig:ev_no_ev_indiv} shows the load profiles at various quantiles of households with or without electric vehicles whilst figure \ref{fig:ev_no_ev_indiv} shows individual samples of load profiles.

\subsection{Utility}

Utility looks at how useful generated samples are in real life applications. In the RCGAN paper, \cite{rcgan} proposes the “Train on Synthetic, Test on Real” (TSTR) framework where two competing models for an evaluation task are trained: one on synthetic data and one on real data. The intuition is that if the synthetic data is useful, then a model trained on synthetic data should perform as well as the model trained with real data. 

\begin{figure}[ht]
\centering
\begin{subfigure}{0.4\textwidth}
\includegraphics[width=0.8\linewidth]{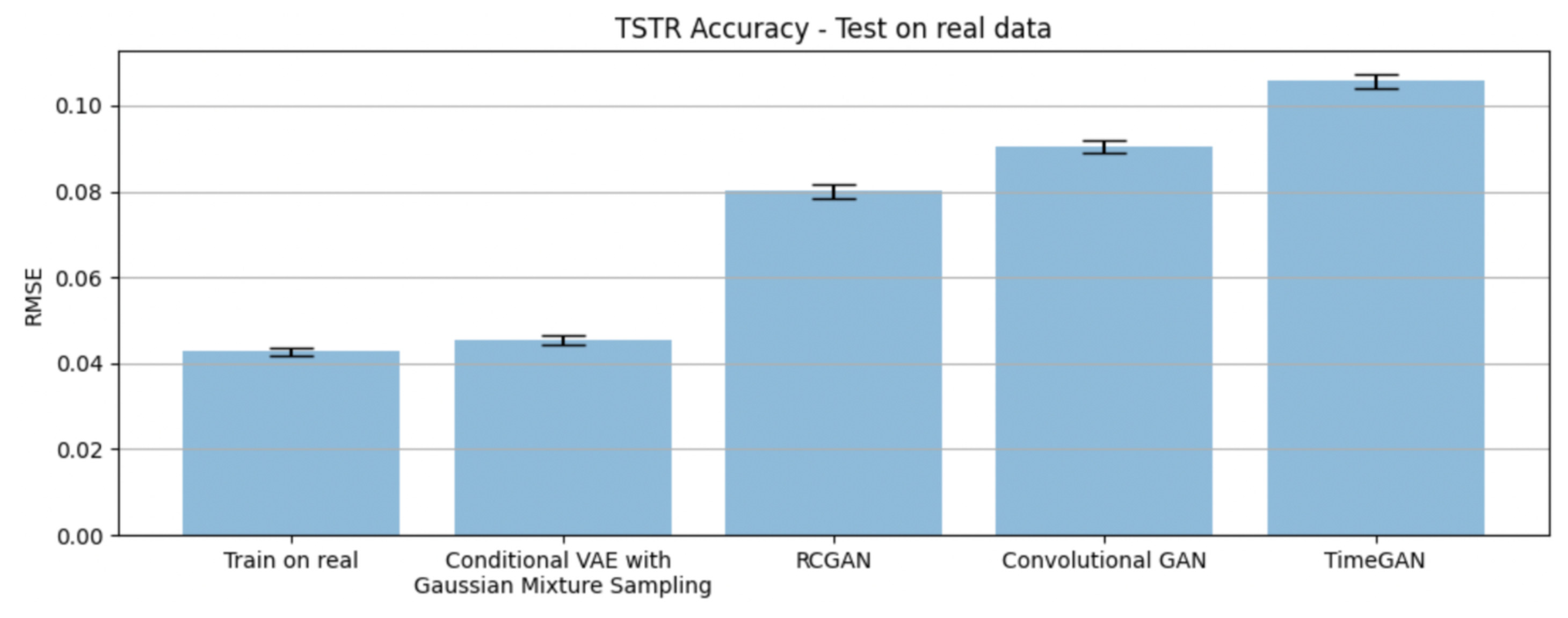}
\caption{TSTR performance.}
\label{fig:tstr}
\end{subfigure}
\begin{subfigure}{0.4\textwidth}
\includegraphics[width=0.8\linewidth]{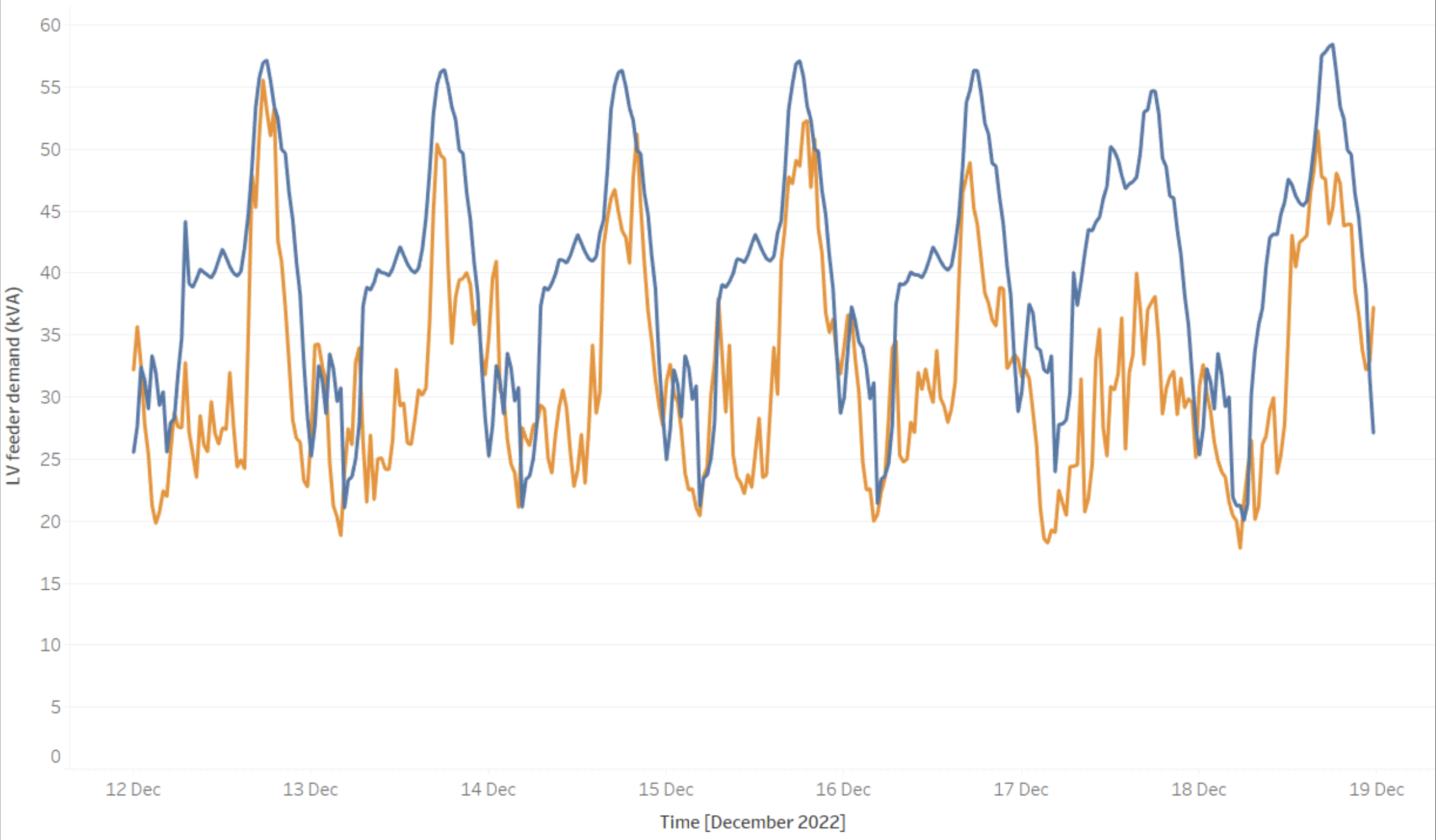}
\caption{Faraday compared to substation data. Blue is Faraday outputs. Orange are substation measured data.}
\label{fig:TEED}
\end{subfigure}

\caption{Utility of Faraday outputs.}
\end{figure}

For this utility task, we used real 2021 consumption data to generate synthetic consumption profiles for 2021. Two forecasting models are then trained, one on real 2021 data and one on synthetic 2021 data, to predict 2022 consumption. Figure \ref{fig:tstr} shows the performance of Faraday outputs is similar to the forecasting model trained on real data. We also compared the utility of Faraday outputs against other popular GAN-based models and show that Faraday outputs have higher utility in this forecasting task.

Faraday outputs were also compared against real-world data measured at substations. \cite{bham_teed} used Faraday outputs in a digital twin project in Birmingham, UK, and compared Faraday outputs to substation measured data and found strong alignment between the two. This was done by collecting property information of houses served by the substation, inputting this information into Faraday to produce household-level profiles and aggregating the profiles to the substation level. Results show that Faraday outputs have similar peaks in terms of magnitude and time, but have higher ‘base load’. This could be due to a skew in population between this dataset and the average UK population (e.g. in terms of income and demographics etc).

\subsection{Privacy}

Privacy refers to the risk of leaking private data because of the model overfitting on the training data. By design, Faraday samples from a distribution to generate synthetic samples during inference and its outputs are therefore ‘synthetic’. However, there could still be a risk of the distribution being overfitted and thus leaking private data, especially in the case of outliers \citep{van_breugel_membership_2023}. Some implicit measures have been implemented to mitigate privacy risks, such as:

\begin{enumerate}
    \item {Applying a k-anonymity of 3 such that at the finest granularity of dimensions there are still at least three households.}
    \item {Exposing Faraday only in a partial black-box setting; users submit inputs via an API to generate outputs. No outputs are generated during sampling if the households matching users' inputs represent a small proportion of the training data.}
    \item {Only outputting daily profiles (as opposed to weekly or monthly profiles) which limits re-identification risk.}
\end{enumerate}

There is a trade-off between privacy and utility. Because of the inherent privacy risks such as the potential re-identification of individuals, Faraday only outputs daily profiles - hence limiting the utility of its data to ‘intraday’ analysis at household level. However, outputs are still useful for inter-day purposes on an aggregated or population level as seen from the comparison to substation outputs in figure \ref{fig:TEED}.

\section{Future work and Conclusion}

Privacy is a central concern when it comes to sharing smart meter data as it contains highly sensitive information. Existing literature on generating synthetic smart meter data places emphasis on fidelity and utility metrics, but there is limited commentary on the privacy of synthetic smart meter data. Whilst Faraday has implicit measures implemented to guard against privacy risks, more work should be undertaken to explicitly quantify the privacy risks of generating synthetic smart meter data. This could include implementing differential privacy \citep{abadi_deep_2016}, or explicit evaluation tasks such as membership inference or reconstruction attacks. Doing so could give us the confidence to output time series of longer horizons, such as weekly or monthly, which would have even higher utility for grid research.

\subsubsection*{Model availability}

As Faraday is trained on proprietary data, the model is only available as a partial black-box access via a web-app and an API. The model is still in closed alpha phase, but we welcome requests for research purposes. For access to Faraday, please contact faraday@centrefornetzero.org with your use case.

\bibliography{iclr2024_conference}
\bibliographystyle{iclr2024_conference}


\end{document}